\documentclass[preprint,3p,times,twocolumn]{elsarticle}

\usepackage{amssymb}
\usepackage{amsmath}
\usepackage{algorithmic}
\usepackage{algorithm}
\usepackage{array}
\usepackage[caption=false,font=normalsize,labelfont=sf,textfont=sf]{subfig}
\usepackage{textcomp}
\usepackage{stfloats}
\usepackage{url}
\usepackage{verbatim}
\usepackage{graphicx}
\usepackage{framed,multirow,color}
\usepackage{threeparttable}
\usepackage{makecell}
\usepackage{balance}
\usepackage{cuted}

\usepackage{etoolbox}
\makeatletter
\patchcmd{\elsarticleprelims}
  {\unvbox\elsarticlehighlightsbox}
  {\begingroup\raggedbottom\unvbox\elsarticlehighlightsbox\endgroup}{}{}
\patchcmd{\elsarticleprelims}
  {\unvbox\elsarticlegrabsbox}
  {\begingroup\raggedbottom\unvbox\elsarticlegrabsbox\endgroup}{}{}
\makeatother

\journal{Biomedical Signal Processing and Control}

\begin{document}

\begin{frontmatter}



\title{RaffeSDG: Random Frequency Filtering enabled Single-source Domain Generalization for Medical Image Segmentation}



\author[label1]{Heng Li\fnref{equal}}
\author[label2]{Haojin Li\fnref{equal}}
\author[label2]{Jianyu Chen}
\author[label2]{Mingyang Ou}
\author[label3]{Hai Shu}
\author[label4,label5]{Heng Miao}
\ead{sawyer_young@sina.com}

\affiliation[label1]{organization={Faculty of Biomedical Engineering, Shenzhen University of Advanced Technology}, 
            city={Shenzhen}, 
            state={Guangdong}, 
            country={China}} 

\affiliation[label2]{organization={The Department of Computer Science and Engineering, Southern University of Science and Technology}, 
            city={Shenzhen}, 
            state={Guangdong}, 
            country={China}} 

\affiliation[label3]{organization={The Department of Biostatistics, School of Global Public Health, New York University}, 
            city={New York}, 
            state={New York}, 
            country={USA}} 

\affiliation[label4]{organization={Department of Ophthalmology, Peking University People's Hospital}, 
            city={Beijing}, 
            country={China}}

\affiliation[label5]{organization={Beijing Key Laboratory of Ocular Disease and Optometry Science, Peking University People’s Hospital}, 
            city={Beijing}, 
            country={China}} 

\fntext[equal]{These authors contributed equally to this work.}

\begin{abstract}
Deep learning models often encounter challenges in making accurate inferences when there are domain shifts between the source and target data. This issue is particularly pronounced in clinical settings due to the scarcity of annotated data resulting from the professional and private nature of medical data. Although various cross-domain strategies have been explored, including frequency-based approaches that vary appearance while preserving semantics, many remain limited by data constraints and computational cost.
To tackle domain shifts in data-scarce medical scenarios, we propose a \textit{Ra}ndom \textit{f}requency \textit{f}iltering \textit{e}nabled \textit{S}ingle-source \textit{D}omain \textit{G}eneralization algorithm (RaffeSDG), which promises robust out-of-domain inference with segmentation models trained on a single-source domain.
A frequency filter-based data augmentation strategy is first proposed to promote domain variability within a single-source domain by introducing variations in frequency space and blending homologous samples.
Then Gaussian filter-based structural saliency is also leveraged to learn robust representations across augmented samples, further facilitating the training of generalizable segmentation models.
To validate the effectiveness of RaffeSDG, we conducted extensive experiments involving out-of-domain inference on segmentation tasks for three human tissues imaged by four diverse modalities. Through thorough investigations and comparisons, compelling evidence was observed in these experiments, demonstrating the potential and generalizability of RaffeSDG.
The code is available at https://github.com/liamheng/Non-IID\_Medical\_Image\_Segmentation.
\end{abstract}




\end{frontmatter}


\section{Introduction}
Deep learning algorithms have garnered widespread recognition in medical research. However, their performance often deteriorates when transitioning from controlled laboratory settings to clinical scenarios due to domain shifts—discrepancies between training (source) and test (target) data distributions caused by differences in imaging modalities, acquisition protocols, or patient demographics~\cite{zhou2022domain,peng2022out}. To mitigate the impact of domain shifts, numerous studies have explored techniques such as domain adaptation (DA) and domain generalization (DG)~\cite{zhou2022domain}. DA focuses on adapting a model trained on a source domain to a target domain, but requires access to the target domain during training, which is often impractical in open clinical datasets. DG addresses this limitation by leveraging multiple source domains to build models that generalize well on unknown domains. Despite their benefits, both DA and DG raise concerns regarding data collection and privacy, especially in clinical scenarios. Consequently, single-source domain generalization (SDG)~\cite{peng2022out} has recently been introduced to enable out-of-domain inference using only a single-source domain, thereby circumventing these concerns.


\begin{figure}[tp]
\centering
\includegraphics[width=1\linewidth]{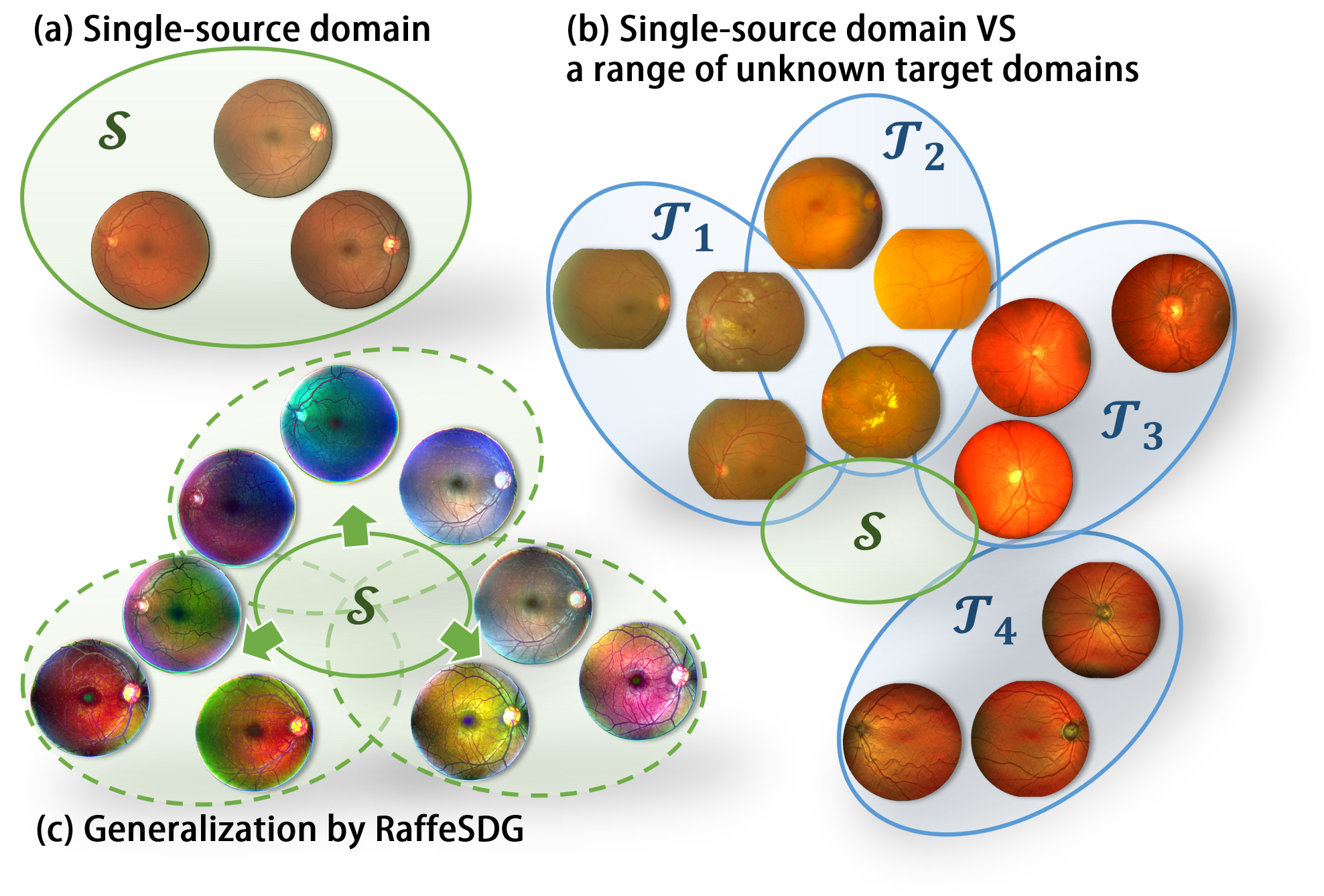}
\caption{
Illustration of single-source domain generalization.
 (c) RaffeSDG enables generalization within (a) a single-source domain,  allowing for inference over (b) a range of unknown target domains.
}
\label{fig:intro}
\vskip -5pt
\end{figure}

As a worst-case scenario of DG (shown in Fig.~\ref{fig:intro} (a) (b)), SDG typically employs augmentation techniques to enhance diversity within the single domain, such as perturbing local textures~\cite{ouyang2022causality,hu2024domain,liang2024single} or transforming global styles~\cite{su2022rethinking} to simulate potential domain variations. While these methods primarily operate in the spatial domain, recent studies have increasingly explored alternative augmentation strategies that manipulate the frequency space. Frequency-based approaches have been widely adopted in cross-domain settings for their ability to induce substantial appearance variations while preserving semantic content~\cite{cheng2023frequency}, making them particularly appealing for scenarios where semantic integrity must be maintained despite distributional shifts. For instance, Yang et al.~\cite{yang2020fda} proposed a frequency-driven method by replacing the amplitude spectrum of a source image with that of a target image. Extending this idea, subsequent algorithms~\cite{xu2021fourier,liu2021feddg} encouraged feature invariance across domains undergoing amplitude spectrum transformations.
More recently, previous work~\cite{li2023frequency} introduced a frequency-oriented SDG paradigm, which leverages parameterized Gaussian filters to manipulate the frequency components of source images for enhancing intra-domain diversity.

Despite advancements in generalizing deep learning models for clinical applications, challenges persist in implementing SDG, particularly the frequency-enabled paradigm.
i) Pioneer SDG algorithms primarily rely on image transformation~\cite{cugu2022attention} or generative networks~\cite{peng2022out, guo2025everything, ouyang2022causality} to enhance data diversity. However, the former may offer negative augmentation, while the latter can escalate the risk of mode collapse.
ii) Though previous frequency-based SDG algorithm alleviated these issues, its use of straightforward Gaussian filters is insufficient to fully diversify the source domain, necessitating more efficient frequency augmentation methods.
iii) Beyond enhancing data diversity within the single-source domain, reasonably leveraging the augmented data to learn and embed generalizable representations is also significant in SDG algorithms.

To facilitate the implementation of SDG, we introduce a 
\textit{\textbf{Ra}}ndom \textit{\textbf{f}}requency 
\textit{\textbf{f}}iltering \textit{\textbf{e}}nabled \textit{\textbf{SDG}} algorithm (RaffeSDG), which allows generalizable segmentation models from a single-source domain without the drawbacks of negative augmentation and mode collapse.
RaffeSDG employs random Fourier filtering and homologous sample blending to augment data diversity within the single-source domain (Fig.~\ref{fig:intro} (c)).
Furthermore, RaffeSDG leverages structure consistency and saliency constraints to impose model robustness and generalizability.
Three human tissues imaged by four modalities are leveraged in the experiment to demonstrate the efficacy of RaffeSDG in out-of-domain inference.
Our main contributions are summarised as follows:
\begin{itemize}
    \item A frequency-enabled SDG paradigm for medical image segmentation termed RaffeSDG is developed, to alleviate the challenges in out-of-domain inference while avoiding the risk of negative augmentation and mode collapse.
    \item A frequency-based augmentation strategy is developed using random Fourier filters and sample blending to attain convenient and effective data diversity within a single-source domain.
    \item Cooperating with the augmentation strategy, constraints on structure consistency and saliency are integrated to boost the robustness and generalizability during model training.
    \item The experiment presents extensive investigations and comparisons using four medical image modalities to demonstrate the benefits and versatility of RaffeSDG in out-of-domain inference.
\end{itemize}

\section{Related Work}
\subsection{Domain adaptation and domain generalization}
Domain shifts severely degrade the performance of deep learning models on cross-domain inference. 
Extensive efforts have been dedicated to alleviating domain shifts, which can be categorized into two approaches: DA and DG.

DA encompasses both supervised and unsupervised paradigms. In supervised DA, the focus is on few-shot adaptation where only a few labeled target samples are involved in training. 
Nevertheless, in medical scenarios, the scarcity of annotated data makes unsupervised DA more prevalent, which aims to align the distribution of the target domain with that of the source domain by minimizing the maximal mean variability or utilizing a domain classifier. 
To achieve adaptation across domains, DAMAN~\cite{mukherjee2022domain} employs a domain regularizing loss to align intermediate features between the source and target domains.
CSCADA~\cite{gu2022contrastive} achieves semi-supervised DA utilizing source data through anatomical structure discrimination.
AIF-SFDA~\cite{li2025aif} trains adaptive filters on unlabeled target data to disentangle domain-specific information and boost source model generalizability.
Nevertheless, accessing the target domain poses a strong requirement in practice, and both paradigms suffer from a strong coupling between the source and target domains, limiting their practicality and generalizability in clinical settings.

DG serves as an alternative pipeline to tackle domain shifts~\cite{zhou2022domain, li2025federated}. In contrast to DA, the objective of DG is to learn from multiple source domains without requiring access to target domains.
This reduced reliance on target data makes DG more appreciated in medical scenarios.
Recently, frequency operations have gained prominence in DG algorithms to generate diverse additional training data by domain randomization~\cite{niu2025exploring}.
FACT~\cite{xu2021fourier} randomizes source domains by swapping the LFS between multiple source domains. Moreover, FedDG~\cite{liu2021feddg} improved the augmentation by employing continuous frequency space interpolation, further enhancing the realism of the generated data.
Apart from augmentation techniques, learning techniques such as feature regularization and self-supervision have gained significant popularity for imposing the learning of domain-invariant representations.
To learn generalizable representations, co-teacher regularization and contrastive learning were respectively leveraged in FACT~\cite{xu2021fourier} and  FedDG~\cite{liu2021feddg}.
Unfortunately, the requirement for multiple distinct source domains in typical DG approaches still poses difficulties in terms of data collection and preserving privacy, especially in clinical settings.

\subsection{Single-source domain generalization}
Due to the data collection burden and privacy concerns in medical scenarios, SDG has become increasingly significant, which involves training a deep model on a single-domain dataset and expecting it to perform well on any unseen domains.

A common SDG strategy is to augment the data distribution while preserving semantic integrity, typically assuming a separation between domain-relevant (e.g., style, texture) and domain-irrelevant (e.g., content) factors.
To this end, commonly adopted techniques include perturbing texture information via randomized auxiliary networks~\cite{ouyang2022causality,hu2024domain,xu2025adversarial}, modulating image style through histogram-based transformations~\cite{su2022rethinking} or randomized normalization methods~\cite{liang2024single} based on adaptive instance normalization (AdaIN)~\cite{huang2017arbitrary}, and applying randomized frequency-domain filtering, which enables effective augmentation without access to target domain spectral information~\cite{li2023frequency}.
Feature-level constraint ~\cite{liang2024single, cho2025peer} is another prevalent strategy in SDG, aiming to improve generalization by regulating the structure of learned representations.
Typical practices include applying contrastive learning to emphasize inter-domain discrepancies between the source and its augmented variants~\cite{liang2024single}, enforcing Lipschitz continuity to reduce sensitivity to domain-specific perturbations~\cite{arslan2024single}.

Despite the impressive success of existing SDG algorithms, their reliance on complex feature operations or auxiliary generative networks hinders practical efficiency and versatility.
To address these limitations, FreeSDG~\cite{li2023frequency} has previously proposed a solution. However, the naive frequency operations employed in FreeSDG~\cite{li2023frequency} restrict the flexibility of data augmentation (refer to Fig.~\ref{fig:augcompare}).
Therefore, RaffeSDG further incorporates random Fourier frequency filters to introduce an efficient SDG paradigm for medical image segmentation.

\section{Method}

\begin{figure*}[tp]
\centering
\includegraphics[width=1\linewidth]{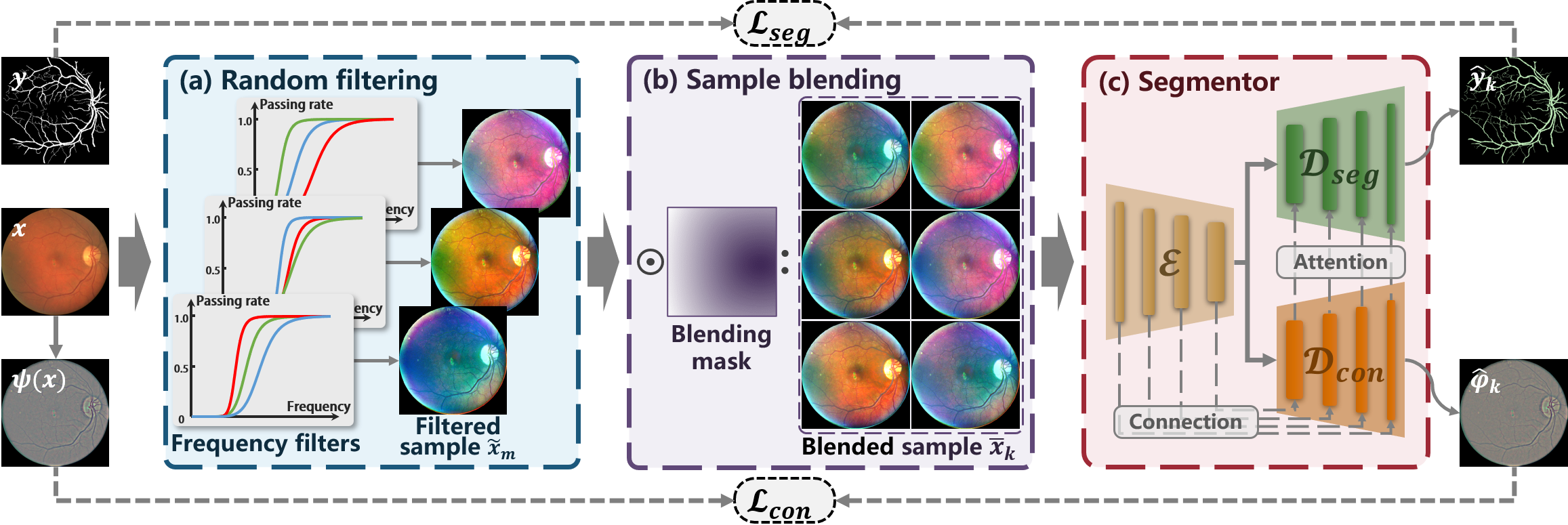}
\caption{
Overview of RaffeSDG. 
Random filtering and sample blending are incorporated to introduce randomization into the single-source domain.
(a) The source image $x$ is first augmented by random high-pass frequency filters.
(b) Sample blending further enables sub-image augmentation by merging filtered samples $\tilde{x}_m$ obtained from $x$. 
(c) The segmentor integrates Gaussian filters to facilitate the learning of domain-invariant representations, and leverages attention mechanisms to appropriately forward the representations for segmentation.
}
\label{fig:workflow}
\vskip -5pt
\end{figure*}

\subsection{Definition and overview}
In DG tasks, the training data from source domains are denoted as $\mathbb{D}_S = \{(x^s_n,y^s_n)^{N_s}_{n=1}\}^{S}_{s=1}$, where $x^s_n$ refers to the $n^{th}$ sample in the $s^{th}$ source domain, $y^s_n$ is the corresponding ground-truth, and $N_s$ and $S$ are the domain volume and the number of source domains, respectively.
Based on $\mathbb{D}_S$, the parameters $\theta$ of a prediction function $g_\theta$ are learned, which maps a sample $x$ from the image space $\mathcal{X}$ to a label $y$ in the label space $\mathcal{Y}$, i.e., $g_\theta: x \to y$.
The main goal of DG is to impose the learned function $g_\theta$ generalizes well to unknown target domains $\mathbb{D}_T = \{(x^t_n)^{N_t}_{n=1}\}^{T}_{t=1}$.
To ensure that $g_\theta$ remains robust against domain shifts between $\mathbb{D}_S$ and $\mathbb{D}_T$, it was common to use multiple source domains ($S>1$) for model training. 
More recently, to address the bottleneck in collecting multiple source domains, SDG has emerged to achieve DG with only one source domain ($S=1$). 

Inspired by the inter-domain attributes of image frequency, we have explored using Gaussian lowpass filters to develop a frequency-enabled SDG paradigm, termed FreeSDG~\cite{li2023frequency}.
Despite the progress achieved by FreeSDG~\cite{li2023frequency}, Gaussian filters suffer a tendency to remove color information, leading to limited diversity in augmentation.
Therefore, we introduce RaffeSDG in this study, which offers flexible data augmentation through random Fourier filtering and homologous sample blending within a single-source domain. 
Subsequently, structure consistency and saliency are incorporated with the augmented data to train generalizable segmentation models.


\subsection{Data augmentation using random frequency filtering} 
Data augmentation has been commonly employed in SDG algorithms to enhance the diversity within a single-source domain.
However, pioneer SDG algorithms employing image transformation~\cite{tiwary2025langdaug, cugu2022attention} or generative networks~\cite{peng2022out, guo2025everything, ouyang2022causality} suffer from negative augmentation or model collapse.
To overcome these limitations, we develop a frequency-based data augmentation strategy using random Fourier filters and sample blending to diversify the single-source domain. 

\subsubsection{Random Fourier filtering with Butterworth filers}
Considering AFS contains low-level image statistics, it has been leveraged in DA and DG to bridge domain shifts. 
Accordingly, we further hypothesize that diversifying a domain can be effectively achieved by adjusting the AFS of image samples. 
To this end, we employ Fourier filtering to enhance the diversity within the single-source domain through data augmentation with AFS adjustments, as illustrated in Fig.~\ref{fig:workflow} (a).



For an image sample $x(a,b)$, it is converted into the frequency domain using Fourier transform: 

\begin{equation}
F(u, v)=\sum_{a=0}^{M-1} \sum_{b=0}^{N-1} x(a, b) e^{-j 2 \pi\left(ua/M+vn/N\right)},
\label{eq:FFT}
\end{equation}
where the image size is $M \times N$. Then a high-pass filter $\eta (u, v)$ is applied to cut off the frequency spectrum, resulting in $G(u, v)=F(u, v)\ast \eta (u, v)$. Finally, the filtered image sample is obtained through inverse Fourier transform:

\begin{equation}
\tilde{x} (a, b)=\frac{1}{MN}\sum_{u=0}^{M-1} \sum_{v=0}^{N-1} G(u, v) e^{j 2 \pi\left(ua/M+vn/N \right)}.
\label{eq:iFFT}
\end{equation}

Additionally, considering filters with sharp transitions can lead to the occurrence of ringing artifacts, the high-pass filter is implemented with Butterworth filters, which is defined as
\begin{equation}
\eta (u, v)=\frac{1}{1+\left[D_{0} / D(u, v)\right]^{2 n}},
\label{eq:Butterworth}
\end{equation}
where $D_{0}$ is the cutoff frequency at which the filter starts attenuating the signal, while $n$ is the order of the filter, influencing the steepness of the filter's roll-off. 

The AFS variations can be introduced by adjusting the parameters of $D_{0}$ and $n$ (the parameter range is interpreted in Sec.~\ref{sec:aug}). Notably, as depicted in Fig.~\ref{fig:workflow} (a), distinct parameters are applied to the filter across different channels, promising the diversity of filtered samples.

To visualize the changes in data distribution resulting from the proposed filtering process, t-SNE is employed in Fig.~\ref{fig:filteraug}. 
This involves dividing images from both the single-source and target domains into patches and embedding them using a pre-trained ResNet. The embedded representations are then mapped into two dimensions using t-SNE, enabling visual observation of the data distribution.
From the comparison between Fig.\ref{fig:filteraug} (a) and (b), it can be observed that random frequency filtering expands the distribution variability within the single-source domain. This serves as evidence that LFS variations can effectively enable domain randomization.

\subsubsection{Homologous sample blending}
Although frequency filtering affects the global image, it lacks the ability to discriminatively alter local areas of the images. To overcome this limitation, we draw inspiration from popular augmentation techniques such as CutOut~\cite{devries2017improved} and CutMix~\cite{yun2019cutmix}. Consequently, homologous filtered samples are blended to introduce additional diversity at the sub-image level, effectively enhancing data augmentation.

Since the filtered samples derived from identical images inherit consistent structures, they can be seamlessly blended together.
Let $\tilde{x}_m$ and $\tilde{x}_n$ represent two filtered samples from the image $x$, where $m,n \in \mathbb{R}^N$ are the sample index and $m\ne n$.
Then the image blending is formulated as:
\begin{equation}
\bar{x}_k= \Xi\left ( \tilde{x}_m,\tilde{x}_n \right ) = \mathcal{M}\odot \tilde{x}_m+(1-\mathcal{M})\odot \tilde{x}_n, 
\label{eq:cutmix}
\end{equation}
where $\mathcal{M}$ is the blending mask and $\odot$ is element-wise multiplication.

As shown in Fig.~\ref{fig:workflow} (b), a continuous blending mask is obtained from a distance map, which is given by:
\begin{equation}
\mathcal{M}(a,b)=\sqrt{(a-c_w)^{2}+(b-c_h)^{2}}/D_{Max},
\label{eq:m3}
\end{equation}
where $(c_w,c_h)$ is the coordinates of the randomly selected distance center within the image, while $D_{Max}$ represents the maximum distance from points in the image to the center.

\begin{figure}[tp]
\centering
\includegraphics[width=1\linewidth]{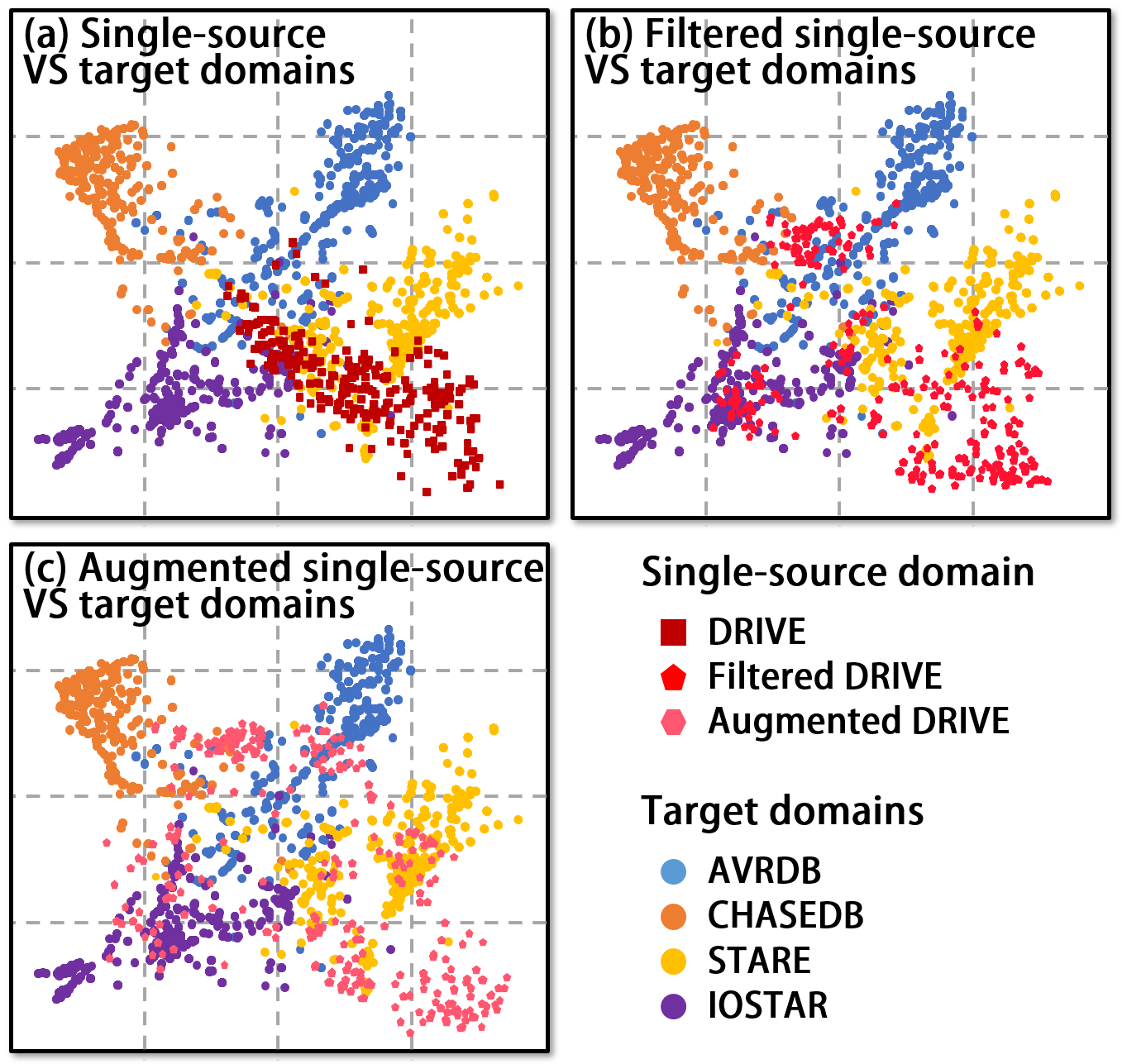}
\caption{
Distributions of the single-source and target domains.
(a) Distributions of patches from original samples in the feature space.
(b) Frequency filtering extended the variability of the single-source domain.
(c) Data augmentation achieved by the proposed strategy.
}
\label{fig:filteraug}
\vskip -5pt
\end{figure}

Consequently, the cooperation of random frequency filtering and homologous sample blending enables sub-image level data augmentation. According to Fig.~\ref{fig:filteraug} (c), this augmentation approach achieves remarkable variability in the single-source domain.

\subsection{Structure consistency and saliency constraints}
The proposed data augmentation promises RaffeSDG a substantial increase in the data diversity within the single-source domain. 
Subsequently, constraints on structure consistency and saliency are introduced to collaborate with the augmented data, further imposing model robustness and generalizability during training.


Given Gaussian filtering allows highlighting structure details~\cite{li2022annotation}, it is incorporated to boost the generalizability of segmentation models by constraining on structure consistency and saliency. 
For an image $x$, the structure saliency map $\psi (x)$ is acquired by: 
\begin{equation}
\psi (x)= x - x\ast g(r,\sigma),
\label{eq:lowpass}
\end{equation}
where $g$ is a Gaussian filter with the kernel of $(r,\sigma)$. 
$r$ and $\sigma$ are determined accordingQ to the image size following~\cite{li2022annotation}. 

Subsequently, to impose constraints on consistency and saliency, the structural saliency map serves as a self-supervised pretext task during model training. The self-supervisory loss is quantified across samples augmented from the identical vanilla images.
The self-supervision loss is defined as: 
\begin{equation}
\mathcal{L}_{sel}=\mathbb{E}\left [{\textstyle\sum_{k=1}^{K}}\left \| \psi (x)-\hat{\varphi}_k \right \|_{2}\right ],
\label{eq:lh}
\end{equation}
where $\hat{\varphi}_k$ refers to the structure saliency map reconstructed from $\bar{x}_k$. 

Specifically, the self-supervised pretext task is designed to reconstruct the structure saliency map from augmented samples.
A coupling segmentation network is built to optimize the model through both self-supervision and segmentation supervision, as illustrated in Fig.~\ref{fig:workflow} (c).
The network consists of an encoder and two decoders.
The encoder $\mathcal{E}$ processes the augmented samples $\bar{x}_k$ and shares its outputs with both the decoder $\mathcal{D}_{sel}$ and $\mathcal{D}_{seg}$, where $\mathcal{D}_{sel}$ predicts the structure saliency and $\mathcal{D}_{seg}$ predicts the segmentation mask.

The concrete implementation of the network is outlined in Algorithm~\ref{alg:training}.
$\mathcal{E}$ is skip-connected to $\mathcal{D}_{sel}$ to facilitate the flow of contextual information, and attention mechanisms are employed to properly integrate the self-supervision into the segmentation task.
The $l^{th}$h layer of the encoder and decoders are denoted as $\mathcal{E}^l$, $\mathcal{D}^l_{sel}$, and $\mathcal{D}^l_{seg}$, with corresponding outcomes $f^{l}_{\mathcal{E}}$, $f^{l}_{sel}$, and $f^{l}_{seg}$. The terms $CA$ and $SA$ represent typical channel and spatial attention respectively~\cite{woo2018cbam}.

\begin{algorithm}[bp]
    \caption{Forward pass in the coupling network}
    \label{alg:training}
    \begin{algorithmic}[1]
        \REQUIRE Augmented views $\{\bar{x}_k|k=1,2,...,K\}$
        \STATE Initialization, $\mathcal{E}$, $\mathcal{D}_{sel}$ and $\mathcal{D}_{seg}$
        \STATE $f^L_{\mathcal{E}}=\mathcal{E}^L(\bar{x}_k)$
        \FOR{$l=L-1,L-2...,1$}
        \STATE $f^l_{\mathcal{\mathcal{E}}}=\mathcal{E}^l(f^{l+1}_{\mathcal{\mathcal{E}}})$        
        \ENDFOR
        \STATE $f^1_{sel}=\mathcal{D}^1_{sel}(f^{1}_{\mathcal{E}})$, $f^1_{seg}=\mathcal{D}^1_{seg}(f^{1}_{\mathcal{E}})$   
        \FOR{$l=2,3,...,L$}
        \STATE $f^l_{sel}=\mathcal{D}^l_{sel}([f^{l-1}_{\mathcal{E}},f^{l-1}_{sel}])$        
        \STATE $f^l_{seg}=\mathcal{D}^l_{seg}([f^{l-1}_{sel},f^{l-1}_{seg}]\otimes CA\otimes SA)$
        \ENDFOR
        \STATE $\hat{\varphi}_k=f^L_{sel}$, $\hat{y}_k=\mathrm{sigmoid}(f^L_{seg})$
        
    \end{algorithmic}
    \vskip -5pt
\end{algorithm}

Given the segmentation objective function by:
\begin{equation}
\mathcal{L}_{seg}=\mathbb{E}\left [- {\textstyle \sum_{c=1}^{C}y^c_k}\log{(\hat{y}^c_k)}\right ],
\label{eq:l_seg}
\end{equation}
where $\hat{y}_k$ denotes the predicted segmentation mask of $\bar{x}_k$, and $c$ is the category index.

Thus the segmentation network is optimized by the overall objective function:
\begin{equation}
\mathcal{L}_{total}=\mathcal{L}_{sel}(\mathcal{E} ,\mathcal{D}_{sel})+\alpha \mathcal{L}_{seg}(\mathcal{E},\mathcal{D}_{sel},\mathcal{D}_{seg}),
\label{eq:overall}
\end{equation}
where $\alpha$ balances $\mathcal{L}_{sel}$ and $\mathcal{L}_{seg}$.

\section{Experiments}
Out-of-domain inference experiments were conducted to demonstrate the potential of RaffeSDG.
The first experiment examined the performance and settings of the proposed augmentation strategy.
Then comprehensive comparisons were conducted between RaffeSDG and SOTA algorithms, including vanilla, DA, DG, and SDG segmentation methods, to present the superiority of RaffeSDG in medical image segmentation tasks.
Additionally, ablation studies were performed to assess the impact of source data volume and the effectiveness of the proposed modules in RaffeSDG.

\subsection{Experiment settings}

\begin{table}[bp]
\footnotesize
\centering 
\caption{Experimental data summary}
\label{tab:data} 
\begin{tabular}{ p{0.9cm} | p{1.3cm}  p{2.6cm}  p{1.4cm}}
\hline
Tissue & Modality & Dataset & Volume\\
\hline
\multirow{3}{*}{Vessel} & Fundus & DRIVE, AVRDB, & 40, 100, 28, \\
& photography & CHASEDB, STARE & 20 \\
\cline{2-4}
& SLO-RGB & IOSTAR  & 30 \\
\hline
Retinal & \multirow{2}{*}{OCT} & HC-MS, DUKE-AMD & 1715, 209, \\
layer &  & GOALS, AROI & 299, 1211 \\
\hline
Joint &  \multirow{2}{*}{Ultrasound} & 4 clinical datasets & 1597, 956, \\
cartilage & & referred to as A$\sim$D & 982, 1455\\
\hline 
\end{tabular}
\end{table}

\subsubsection{Datasets}
The experiments utilized four medical imaging modalities obtained from three different human tissues, as summarized in Table~\ref{tab:data}. 
The data for the vessel and retinal layer were publicly available, while the data for the joint cartilage were collected in collaboration with our partner, Southern University of Science and Technology Hospital, using discrepant ultrasound configurations of gain and grayscale.
In order to generalize and assess segmentation models across these datasets, the ground-truth annotations were standardized.
The datasets of DRIVE, HC-MS, and A were respectively employed as the single-source domain for each tissue, while the remaining datasets were treated as unseen target domains.

Moreover, during the comparisons, the target domains were visible to the DA algorithms, and the extensive fundus photography dataset of EyePACS was incorporated as additional source domains to implement typical DG algorithms.

\subsubsection{Implementation}
For all segmentation tasks, the input image size was set to $512 \times 512$, and a batch size of 2. 
The training data were divided into training and validation subsets in 1:1.
The models were trained using the Adam optimizer. Initially, the learning rate was set to 1e-3 for the first 80 epochs. Subsequently, a linear decrease to zero was applied over the next 120 epochs. 
In each epoch, ten augmented samples were randomly derived from each training sample.
To ensure efficient training, the early stop was implemented. 
During the inference, the segmentation is performed using the original images without any augmentation.

\subsection{Data augmentation configuration in RaffeSDG}
\label{sec:aug}
We conduct an investigation into the data augmentation configuration of RaffeSDG, which encompasses the exploration of filter types and blending strategies, as well as the random range of parameters.

\subsubsection{Filter types and blending strategies}
\begin{figure}[thbp]
\centering
\includegraphics[width=1\linewidth]{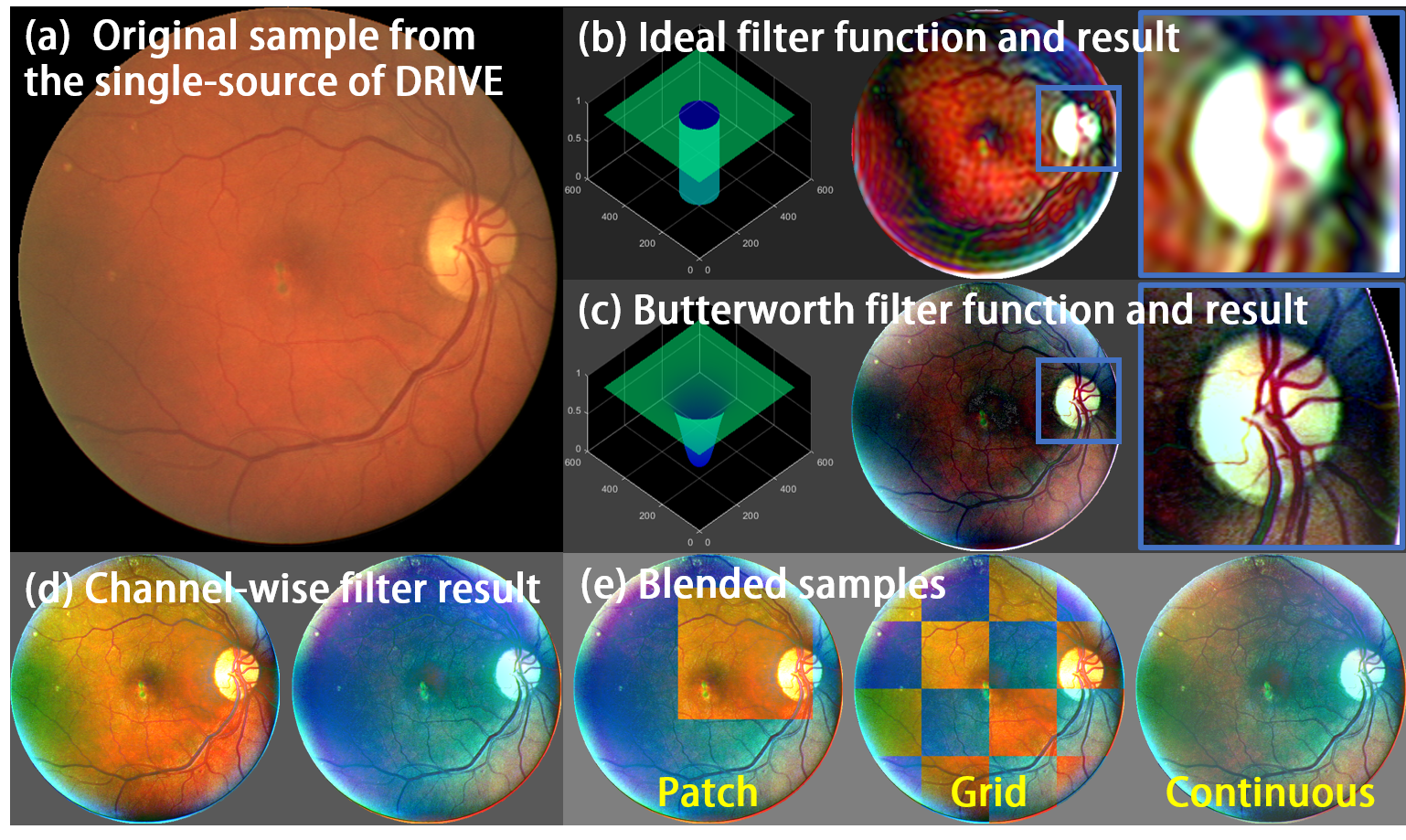}
\caption{
Frequency-filtered images. (a) Original image. (b) Function diagram and filtered result using an ideal filter. (c) Function diagram and filtered image using a Butterworth filter. (d) Filtered images using channel-wise Butterworth filters. (e) Samples blended from filtered ones. 
}
\label{fig:AugParam}
\end{figure}

To validate the configuration of the proposed data augmentation approach in RaffeSDG, we explored different filter types and blending strategies. The filtered and blended samples are illustrated in Fig.~\ref{fig:AugParam}.
Moreover, training data is augmented from DRIVE, and a vanilla U-Net is trained to quantitatively evaluate the performance of various configurations on the target domains, as summarized in Table~\ref{tab:AugParam}.

\noindent \textbf{Filter type:}
To analyze the impact of filter types on data augmentation, we employ both the ideal filter and the Butterworth filter to introduce variations into the LFS of images.
Fig.\ref{fig:AugParam} (b) presents the function diagram and the filtered result of the ideal filter.
The sharp cutoff frequency of the ideal filter leads to the presence of ringing artifacts in the filtered result.
On the other hand, the Butterworth filter offers a smoother transition between the passband and stopband, reducing the occurrence of ringing artifacts. Thus, the Butterworth filter is deemed more suitable for data augmentation, as illustrated in Fig.\ref{fig:AugParam} (c). 
Additionally, to enhance diversity in the augmented samples, we implement a channel-wise filter by applying Butterworth filters with different parameters to the RGB channels, as depicted in Fig.~\ref{fig:AugParam} (d).
Consequently, superior performance is achieved by the channel-wise filter in Table~\ref{tab:AugParam}.

\noindent \textbf{Blending strategy:}
To interpret the effect of blending, we compare three strategies, including patch, grid, and continuous mask, as illustrated in Fig.\ref{fig:AugParam} (e). The images blended using patch and grid masks exhibit a noticeable gap, which deviates from real-world scenarios and hampers the overall context. In contrast, the continuous mask enables a seamless context in the augmented image, resulting in a more coherent representation that aligns well with real-world scenarios. The evaluation results provided in Table\ref{tab:AugParam} further validate our choice of filter and blending configurations.
Accordingly, the channel-wise filter and the continuous mask are cooperated to implement RaffeSDG.

\begin{table}[tp]
\scriptsize
\centering
\caption{Augmentation configuration comparison on training U-Net using only DRIVE for out-of-domain segmentation.}
\label{tab:AugParam}
\renewcommand{\arraystretch}{1.15}
\begin{tabular}{p{1.85cm} | p{0.65cm}<{\centering} p{0.8cm}<{\centering} p{0.65cm}<{\centering}  p{0.65cm}<{\centering} p{0.65cm}<{\centering}}
\hline
\multirow{2}{*}{Configuration}  & \multicolumn{5}{c}{DICE}\\
\cline{2-6}
& AVRDB & CHASEDB & STARE & IOSTAR & Avg.\\ 
\hline
U-Net only & 0.549 & 0.219 & 0.600 & 0.594 & 0.491 \\
\hline
Ideal filter & 0.333 & 0.365 & 0.425 & 0.540 & 0.416\\
Butterworth filter & 0.510 & 0.411 & 0.602 & 0.670 & 0.548\\
Channel-wise filter & \textbf{0.570} & \textbf{0.583} & \textbf{0.671} & \textbf{0.701} & \textbf{0.631}\\
\hline
Patch mask & 0.586 & 0.591 & 0.692 & 0.706 & 0.639\\
Grid mask & 0.580 & 0.586 & 0.685 & 0.703 & 0.646\\
Continuous mask & \textbf{0.593} & \textbf{0.691} & \textbf{0.699} & \textbf{0.720} & \textbf{0.676}\\
\hline
\end{tabular}%
\vskip -10pt
\end{table}

\subsubsection{Parameter range}
\begin{figure}[tbp]
\centering
\includegraphics[width=1\linewidth]{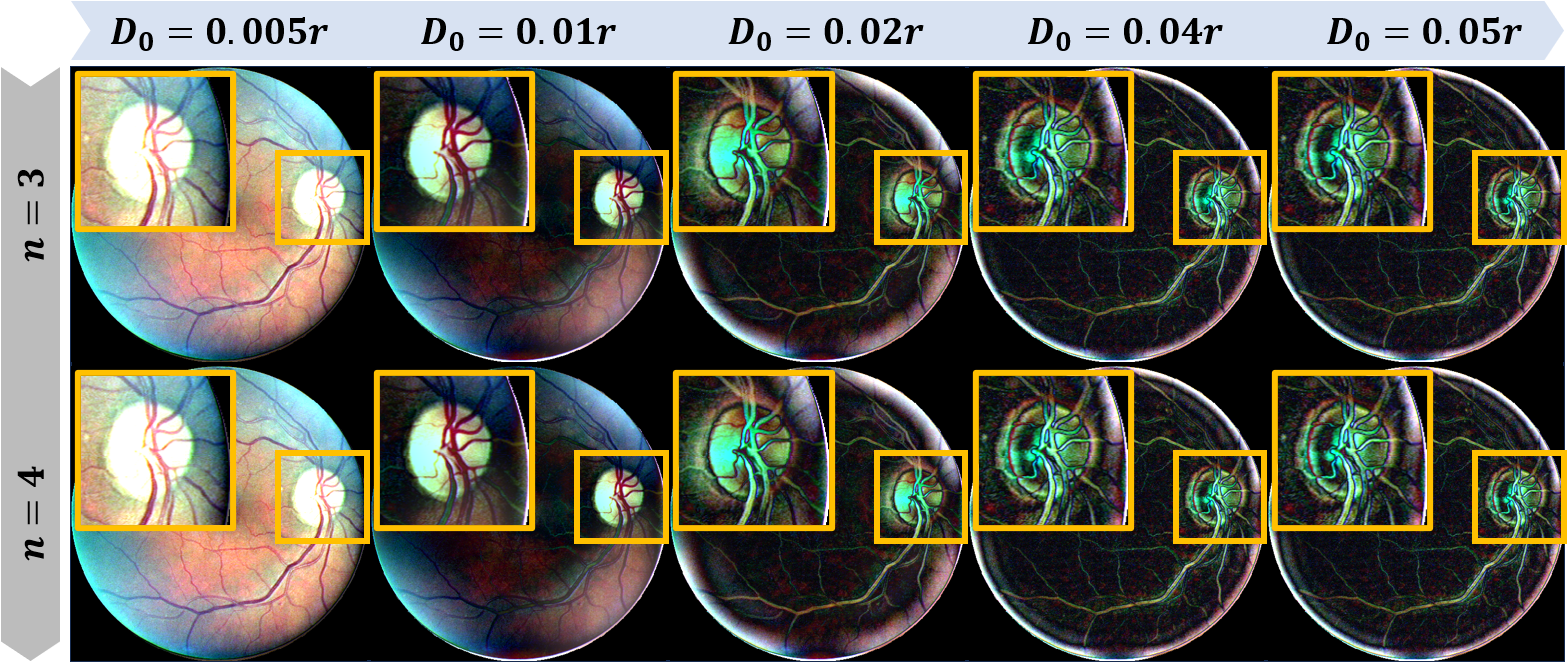}
\caption{
Filtering results with varying $D_0$ and $n$, where $r$ represents the radius of the spectrum of an image.
}
\label{fig:filterparam}
\end{figure}

According to Eq.~\ref{eq:Butterworth}, the Butterworth filter is controlled by two parameters, $D_0$ and $n$. 
$D_{0}$ determines the frequency at which the filter starts attenuating the signal, while $n$ determines the steepness of the filter's roll-off.
Fig.~\ref{fig:filterparam} illustrates the impact of varying $D_0$ and $n$ on the Butterworth filter's behavior.

As shown in Fig.~\ref{fig:filterparam}, $n$ has a minor influence on the filtered results, while increasing the value of $D_0$ leads to a reduction in the low-frequency style and an increase in the visibility of artifacts. However, using high values for both $n$ and $D_0$ can aggravate the appearance of ringing artifacts.
Based on these observations, we have determined that setting the parameter range as $D_0 \le 0.04r$ ($r$ is the radius of the spectrum of an image) and $n \le 3$ to provide satisfactory results and mitigate the risk of unwanted artifacts.


\subsection{Augmentation comparison}
Considering the expansion of the single-source domain plays a crucial role in achieving SDG, a comparison focusing on augmentation performance is conducted to validate the benefit of the proposed augmentation in RaffeSDG.

\begin{figure}[t]
\centering
\includegraphics[width=1\linewidth]{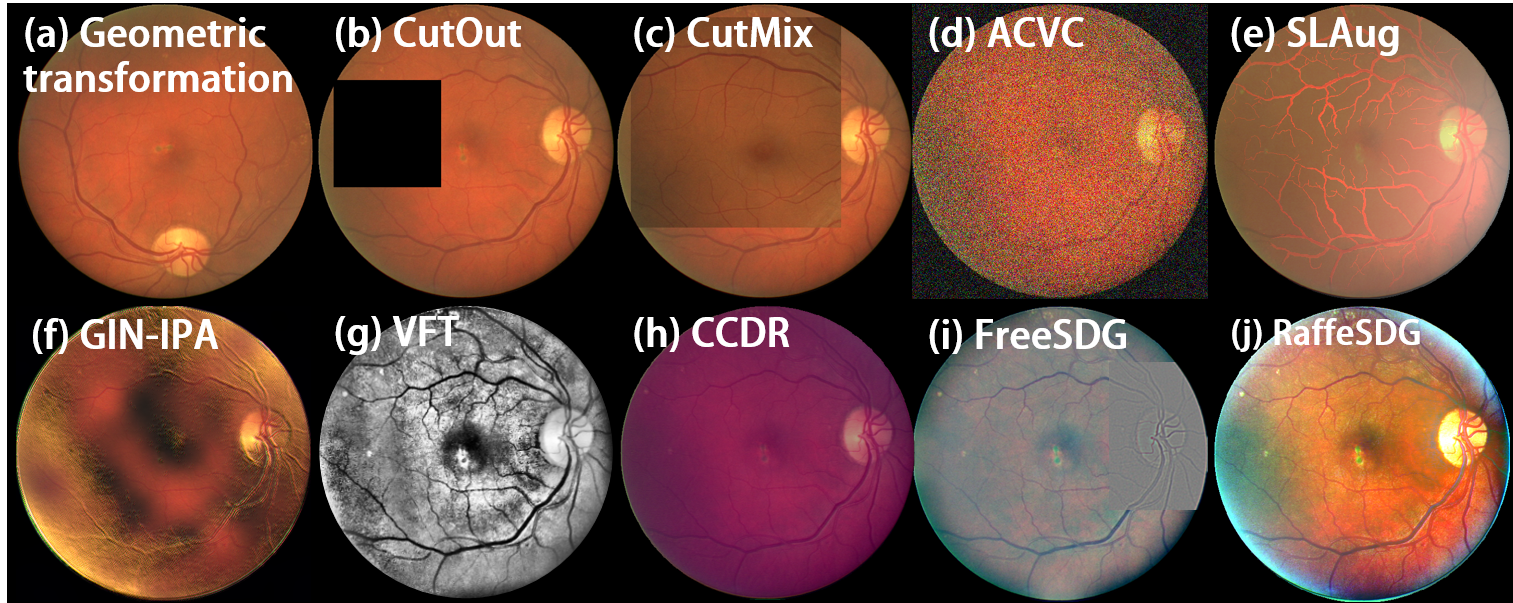}
\caption{
Samples augmented by various augmentation techniques. 
}
\label{fig:augcompare}
\end{figure}

The comparison includes both commonly used augmentation techniques and augmentation approaches designed in SDG algorithms.
Specifically, the augmentation techniques include geometric transformation, CutOut~\cite{devries2017improved}, and CutMix~\cite{yun2019cutmix}.
And the SDG augmentation approaches are collected from ACVC~\cite{cugu2022attention}, SLAug~\cite{su2022rethinking}, GIN-IPA~\cite{ouyang2022causality}, VFT~\cite{hu2024domain}, CCDR~\cite{liang2024single}, and FreeSDG~\cite{li2023frequency}.
In the comparison, the dataset of DRIVE is utilized as the single-source domain, and a vanilla U-Net serves as the unified segmentor, which is trained with augmented data and conducts out-of-domain inference. The augmented samples are visualized in Fig.~\ref{fig:augcompare}, while the segmentation performance is quantified in Table~\ref{tab:augcompare} by training a U-Net.

As depicted in Fig.~\ref{fig:augcompare}, the native augmentation strategy of geometric transformation, CutOut~\cite{devries2017improved}, and CutMix~\cite{yun2019cutmix} are constrained to the single-source domain, leading to limited appearance variations in the augmented samples.
In contrast, the augmentation approaches used in SDG tasks present remarkable performance in Table~\ref{tab:augcompare}. 
For implementing SDG in classification, ACVC~\cite{cugu2022attention} introduces 19 visual corruptions and 3 Fourier-based corruptions to augment the single-source domain.
For medical image segmentation, SLAug~\cite{su2022rethinking} introduces a saliency-balancing fusion strategy that combines global-scale and local location-scale aspects to augment the single-source domain.
GIN-IPA~\cite{ouyang2022causality} proposes a causality-inspired data augmentation approach, which utilizes global intensity non-linear augmentation to diversify the appearances of medical images and addresses confounders through interventional pseudo-correlation augmentation.
VFT~\cite{hu2024domain} performs randomized image reconstruction with cascaded autoencoders, using a dimensionally matched intermediate output as an information-complete augmentation to enrich the source domain.
Moreover, CCDR~\cite{liang2024single} combines random convolution–based texture augmentation and randomized instance normalization–based style augmentation to enrich source domain feature patterns from multiple aspects.

\begin{figure*}[htbp]
\centering
\includegraphics[width=1\linewidth]{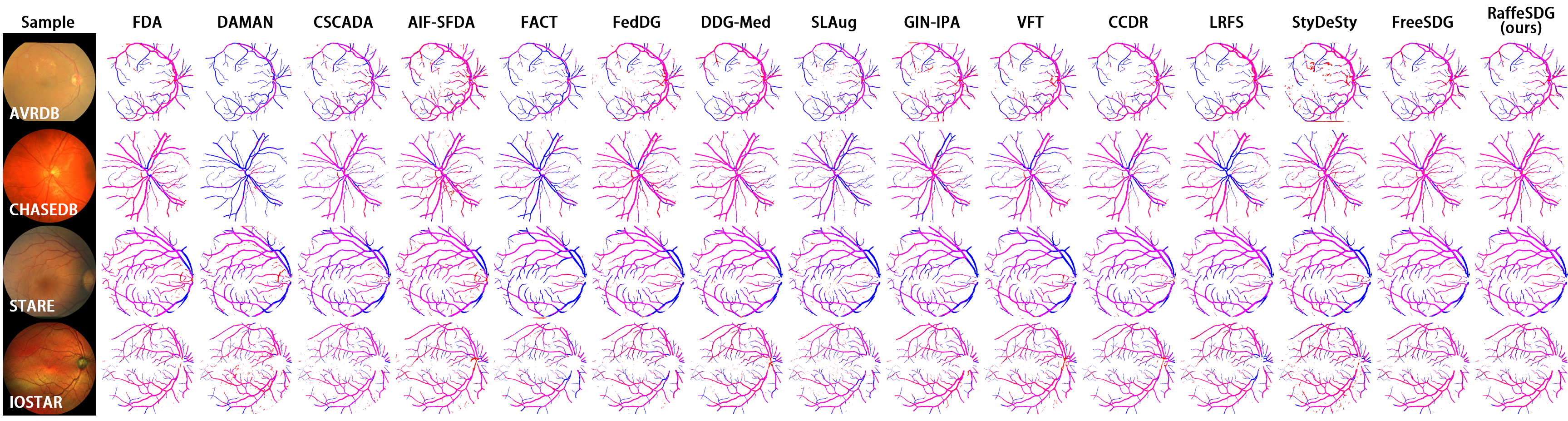}
\caption{
Visualized comparison of out-of-domain inferences among SOTA segmentation algorithms. The \textcolor{blue}{ground-truth masks} are represented in blue, the \textcolor{red}{predictions} made by the algorithms are represented in red, and the \textcolor{magenta}{correctly identified pixels} are represented in magenta.
}
\label{fig:compare1}
\vskip -5pt
\end{figure*}

\begin{table}[tbp]
\scriptsize
\centering
\caption{Augmentation technique comparison on training U-Net. The top three results are indicated in \textbf{\color{red}red}, \textbf{\color{magenta}magenta}, and \textbf{\color{blue}blue}.}
\label{tab:augcompare} 
\renewcommand{\arraystretch}{1.15}
\begin{tabular}{p{1.5cm} | p{0.65cm}<{\centering} p{0.9cm}<{\centering} p{0.65cm}<{\centering}  p{0.65cm}<{\centering} p{0.65cm}<{\centering}}
\hline
\multirow{2}{*}{Augmentation} & \multicolumn{5}{c}{DICE }\\
\cline{2-6}
& AVRDB & CHASEDB & STARE & IOSTAR & Avg.\\ 
\hline
Geometric & 0.549 & 0.219 & 0.600 & 0.594 & 0.491 \\
CutOut~\cite{devries2017improved} & 0.502 & 0.189 & 0.606 & 0.411 & 0.427 \\
CutMix~\cite{yun2019cutmix} & 0.569 & 0.390 & 0.609 & 0.665 & 0.558 \\
ACVC~\cite{cugu2022attention} & 0.609 & 0.613 & \textbf{\color{red}0.727} & \textbf{\color{magenta}0.701} & \textbf{\color{blue}0.663}\\
SLAug~\cite{su2022rethinking} & 0.569 & 0.580 & 0.635 & 0.646 & 0.608 \\
GIN-IPA~\cite{ouyang2022causality} & 0.580 & \textbf{\color{blue}0.647} & 0.674 & \textbf{\color{blue}0.694} & 0.649\\
VFT~\cite{hu2024domain} & \textbf{\color{red}0.634} & 0.636 & 0.685 & 0.685 & 0.660 \\
CCDR~\cite{liang2024single} & \textbf{\color{blue}0.610} & 0.642 & \textbf{\color{magenta}0.701} & 0.663 & 0.654 \\
FreeSDG~\cite{li2023frequency} & \textbf{\color{magenta}0.632} & \textbf{\color{magenta}0.671} & 0.677 & 0.690 & \textbf{\color{magenta}0.668}\\
FreeAug (ours) & 0.593 & \textbf{\color{red}0.691} & \textbf{\color{blue}0.699} & \textbf{\color{red}0.720} & \textbf{\color{red}0.676}\\
\hline
\end{tabular}%
\vskip -5pt
\end{table}

FreeSDG~\cite{li2023frequency} utilized Gaussian filters and patch masks to augment the single-source domain. However, in Fig.~\ref{fig:augcompare} (f), limited diversity was observed in the augmentation performed by FreeSDG~\cite{li2023frequency}, resulting in suboptimal performance as indicated in Table~\ref{tab:augcompare}.
Fortunately, the augmentation approach proposed in RaffeSDG successfully introduces a wide range of visual variations in the single-source domain, as illustrated in Fig.~\ref{fig:augcompare} (h). This diversification strategy has proven to be effective, surpassing alternative methods in training the U-Net model and achieving superior performance, as demonstrated in Table~\ref{tab:augcompare}.

\begin{table*}[!t]
\scriptsize
\centering
\caption{Comparison with SOTA segmentation algorithms. The top three results are indicated in \textbf{\color{red}red}, \textbf{\color{magenta}magenta}, and \textbf{\color{blue}blue}.}
\label{tab:comparison_fundus} 
\renewcommand{\arraystretch}{1.15}
\begin{tabular}{p{1.8cm}| p{0.3cm}<{\centering} p{0.3cm}<{\centering}| p{0.8cm}<{\centering} p{0.9cm}<{\centering} p{0.8cm}<{\centering}  p{0.8cm}<{\centering} p{0.8cm}<{\centering} |p{0.8cm}<{\centering} p{0.9cm}<{\centering} p{0.8cm}<{\centering}  p{0.8cm}<{\centering} p{0.8cm}<{\centering}}
\hline
\multirow{2}{*}{Algorithms}  & \multicolumn{2}{c|}{Dependency*} & \multicolumn{5}{c|}{DICE} &  \multicolumn{5}{c}{IoU}\\
\cline{2-13}
& \textsl{VT} & \textsl{MS} & AVRDB & CHASEDB & STARE & IOSTAR & Avg. & AVRDB & CHASEDB & STARE & IOSTAR & Avg.\\ 
\hline
CS-Net~\cite{mou2019cs} & / & / & 0.576 & 0.233 & 0.580 & 0.576 & 0.491 & 0.324 & 0.132 & 0.408 & 0.405 & 0.317 \\
UTNet~\cite{gao2021utnet} & / & / & 0.527 & 0.410 & 0.661 & 0.614 & 0.553 & 0.357 & 0.258 & 0.493 & 0.443 & 0.388 \\
Dtmformer~\cite{wang2024dtmformer} & / & / & 0.618 & 0.628 & 0.714 & 0.683 & 0.661 & 0.484 & 0.458 & 0.573 & 0.520 & 0.501 \\
Rolling-unet~\cite{liu2024rolling} & / & / & 0.622 & 0.598 & 0.713 & 0.651 & 0.646 & 0.456 & 0.421 & 0.567 & 0.483 & 0.482 \\
FDA~\cite{yang2020fda} & \textcolor{red}{$\star$} & & 0.660 & 0.683 & \textbf{\color{blue}0.759} & \textbf{\color{blue}0.720} & \textbf{\color{magenta}0.705} & 0.495 & 0.514 & \textbf{\color{blue}0.618} & \textbf{\color{blue}0.564} & \textbf{\color{magenta}0.548} \\
DAMAN~\cite{mukherjee2022domain} & \textcolor{red}{$\star$} & & 0.508 & 0.395 & 0.689 & 0.682 & 0.569 & 0.341 & 0.228 & 0.519 & 0.517 & 0.401 \\
CSCADA~\cite{gu2022contrastive} & \textcolor{red}{$\star$} & &  0.630 & \textbf{\color{magenta}0.694} & 0.703 & \textbf{\color{magenta}0.741} & 0.692 & 0.460 & \textbf{\color{magenta}0.531} & 0.542 & \textbf{\color{magenta}0.588} & 0.530 \\
AIF-SFDA~\cite{li2025aif} & \textcolor{red}{$\star$} & & \textbf{\color{red}0.670} & 0.648 & \textbf{\color{magenta}0.762} & 0.719 & \textbf{\color{blue}0.700} & \textbf{\color{red}0.508} & 0.481 & \textbf{\color{magenta}0.622} & 0.562 & \textbf{\color{blue}0.543} \\
FACT~\cite{xu2021fourier} & & \textcolor{red}{$\star$} & 0.581 & 0.480 & 0.561 & 0.661 & 0.571 & 0.409 & 0.325 & 0.390 & 0.493 & 0.404 \\
FedDG~\cite{liu2021feddg} & & \textcolor{red}{$\star$} & \textbf{\color{magenta}0.663} & 0.659 & 0.752 & 0.697 & 0.693 & \textbf{\color{magenta}0.496} & \textbf{\color{blue}0.492} & 0.602 & 0.535 & 0.531 \\
DDG-Med~\cite{cheng2025dynamic} & & \textcolor{red}{$\star$} & 0.661 & \textbf{\color{blue}0.668} & 0.751 & 0.672 & 0.688 & \textbf{\color{blue}0.495} & 0.503 & 0.608 & 0.507 & 0.528 \\
RaffeSDG (ours) && & \textbf{\color{blue}0.661} & \textbf{\color{red}0.731} & \textbf{\color{red}0.775} & \textbf{\color{red}0.749} & \textbf{\color{red}0.729} & 0.493 & \textbf{\color{red}0.576} & \textbf{\color{red}0.633} & \textbf{\color{red}0.599} & \textbf{\color{red}0.575} \\
\hline
\end{tabular}%
\begin{tablenotes}
 \scriptsize
 \item{*} Dependency on visiting test data (\textsl{VT}) and multiple source domains  (\textsl{MS}) are indicated by \textcolor{red}{$\star$}. 
\end{tablenotes}
\vskip -10pt
\end{table*}

\subsection{Comparison with segmentation algorithms}
Comparisons with SOTA segmentation and SDG algorithms are conducted to validate the advantages of the proposed RaffeSDG.
The visualized results are presented in Fig.~\ref{fig:compare1}, while quantitative evaluations are summarized in Table~\ref{tab:comparison_fundus} and~\ref{tab:comparison_SDG}.

\subsubsection{Comparison with segmentation algorithms}
We employ 11 SOTA segmentation algorithms as baselines.
Specifically, CS-Net~\cite{mou2019cs}, UTNet~\cite{gao2021utnet}, Dtmformer~\cite{wang2024dtmformer} and Rolling-unet~\cite{liu2024rolling} are utilized, which are vanilla segmentation algorithms for medical images trained solely on the training data. For domain adaptive segmentation algorithms, FDA~\cite{yang2020fda}, DAMAN~\cite{mukherjee2022domain}, CSCADA~\cite{gu2022contrastive} and AIF-SFDA~\cite{li2025aif} are implemented, which are trained using both the training data and unlabeled test data. Additionally, FACT~\cite{xu2021fourier}, FedDG~\cite{liu2021feddg} and DDG-Med~\cite{cheng2025dynamic} are also included, which are segmentation algorithms based on domain generalization trained with multiple source domains.

The comparative algorithms are trained based on their data dependency, with the DRIVE dataset serving as the default training data. The segmentation results are summarized in Table~\ref{tab:comparison_fundus}.
Due to the lack of specific design for handling domain shifts, CS-Net~\cite{mou2019cs}, UTNet~\cite{gao2021utnet}, Dtmformer~\cite{wang2024dtmformer} and Rolling-unet~\cite{liu2024rolling} trained on the DRIVE dataset experience a notable decline in performance when applied to other datasets.
By incorporating unlabeled test data during model training, FDA~\cite{yang2020fda}, DAMAN~\cite{mukherjee2022domain}, CSCADA~\cite{gu2022contrastive} and AIF-SFDA~\cite{li2025aif} bridge domain shifts, resulting in improved performance in the target domains.
Furthermore, FACT~\cite{xu2021fourier}, FedDG~\cite{liu2021feddg} and DDG-Med~\cite{cheng2025dynamic} utilize multiple source domains to learn generalizable segmentation models, demonstrating decent performance on unseen target domains.

Despite relying solely on the single-source domain of DRIVE without any additional data, the proposed RaffeSDG effectively learns generalizable segmentation models, leading to superior performance on newly encountered datasets.

\begin{table*}[!t]
\scriptsize
\centering
\caption{Comparison with SDG algorithms.}
\label{tab:comparison_SDG} 
\renewcommand{\arraystretch}{1.15}
\begin{tabular}{p{1.8cm} | p{0.9cm}<{\centering} p{0.9cm}<{\centering} p{0.9cm}<{\centering}  p{0.9cm}<{\centering} p{0.9cm}<{\centering} |p{0.9cm}<{\centering} p{0.9cm}<{\centering} p{0.9cm}<{\centering}  p{0.9cm}<{\centering} p{0.9cm}<{\centering}}
\hline
\multirow{2}{*}{SDG algorithms} &  \multicolumn{5}{c|}{DICE} &  \multicolumn{5}{c}{IoU}\\
\cline{2-11}
& AVRDB & CHASEDB & STARE & IOSTAR & Avg. & AVRDB & CHASEDB & STARE & IOSTAR & Avg.\\ 
\hline
SLAug~\cite{su2022rethinking} & 0.626 & 0.662 & 0.717 & 0.683 & 0.672 & 0.456 & 0.495 & 0.559 & 0.519 & 0.507 \\
GIN-IPA~\cite{ouyang2022causality}* & 0.649 & 0.654 & 0.712 & \textbf{\color{blue}0.717} & \textbf{\color{blue}0.683} & 0.481 & 0.486 & 0.553 & \textbf{\color{blue}0.559} & \textbf{\color{blue}0.520} \\
VFT~\cite{hu2024domain} & 0.638 & 0.646 & 0.705 & 0.710 & 0.675 & 0.470 & 0.473 & 0.546 & 0.551 & 0.510 \\
CCDR~\cite{liang2024single} & \textbf{\color{blue}0.652} & \textbf{\color{magenta}0.706} & \textbf{\color{red}0.780} & 0.705 & 0.711 & \textbf{\color{blue}0.484} & \textbf{\color{magenta}0.547} & \textbf{\color{red}0.643} & 0.544 & 0.554 \\
LRFS~\cite{arslan2024single} & 0.625 & 0.621 & 0.638 & 0.646 & 0.632 & 0.464 & 0.459 & 0.476 & 0.483 & 0.471 \\
StyDeSty~\cite{liu2024stydesty} & 0.618 & 0.627 & 0.637 & 0.641 & 0.631 & 0.448 & 0.456 & 0.466 & 0.471 & 0.460 \\
FreeSDG~\cite{li2023frequency} & \textbf{\color{red}0.669} & \textbf{\color{blue}0.696} & \textbf{\color{blue}0.732} & \textbf{\color{magenta}0.736} & \textbf{\color{magenta}0.708} & \textbf{\color{red}0.503} & \textbf{\color{blue}0.533} & \textbf{\color{blue}0.578} & \textbf{\color{magenta}0.582} & \textbf{\color{magenta}0.549}\\
RaffeSDG (ours) & \textbf{\color{magenta}0.661} & \textbf{\color{red}0.731} & \textbf{\color{magenta}0.775} & \textbf{\color{red}0.749} & \textbf{\color{red}0.729} & \textbf{\color{magenta}0.493} & \textbf{\color{red}0.576} & \textbf{\color{magenta}0.633} & \textbf{\color{red}0.599} & \textbf{\color{red}0.575} \\
\hline
\end{tabular}%
\begin{tablenotes}
 \scriptsize
 \item{*} A few cases of failure to converge were also observed, due to model collapse.
\end{tablenotes}
\vskip -10pt
\end{table*}

\subsubsection{Comparison with SDG algorithms}
Seven SDG algorithms, including SLAug~\cite{su2022rethinking}, GIN-IPA~\cite{ouyang2022causality}, VFT~\cite{hu2024domain}, CCDR~\cite{liang2024single}, LRFS~\cite{arslan2024single}, StyDeSty~\cite{liu2024stydesty}, and FreeSDG~\cite{li2023frequency} are also compared and summarized in Table~\ref{tab:comparison_SDG}.
SLAug~\cite{su2022rethinking} introduces a saliency-balancing fusion strategy for data augmentation, which enhances out-of-domain segmentation performance under a single-source domain setting.
GIN-IPA~\cite{ouyang2022causality} employs global intensity non-linear augmentation to diversify image appearances and utilizes interventional pseudo-correlation augmentation to mitigate the influence of confounding factors.
VFT~\cite{hu2024domain} enhances model generalization by integrating Hessian matrix-based vector flow feature extraction with latent space uncertainty-guided image augmentation.
CCDR~\cite{liang2024single} adopts domain randomization in terms of texture and style and performs class-based domain alignment between the source and auxiliary domains to improve adaptation to the target domain.
LRFS~\cite{arslan2024single} enforces a Lipschitz constraint in the image frequency spectrum, reducing the sensitivity to high-frequency components and improving the robustness in the mid-frequency range, thus lowering the reliance on domain-specific information.
Meanwhile, StyDeSty~\cite{liu2024stydesty} utilizes a min-max cooperative optimization framework between stylization and de-stylization modules, enabling the model to learn domain-invariant representations.
Though GIN-IPA~\cite{ouyang2022causality} and CCDR~\cite{liang2024single} demonstrate impressive performance, the use of auxiliary generative networks increases the risk of model collapse.

By incorporating frequency variations and employing a jointly supervised segmentor, both FreeSDG~\cite{li2023frequency} and RaffeSDG demonstrate exceptional performance compared to other SDG paradigms. They also offer the advantage of convenient implementation, eliminating the need for intricate calculations and generative network tuning.
However, due to the limited diversity of frequency variations using Gaussian filters, FreeSDG~\cite{li2023frequency} exhibits suboptimal performance when compared to RaffeSDG.
In contrast, RaffeSDG introduces more flexible frequency variation and image blending strategies, enabling robust randomization within the single-source domain. This leads to superior performance in out-of-domain segmentation tasks.

\subsection{Ablation studies}
\subsubsection{Data volume impact}
To gain a comprehensive understanding of the potential of RaffeSDG, we conducted an analysis to assess the influence of data volume from a single-source. Specifically, we progressively increased the number of raw samples from DRIVE to examine its impact on the generalization performance of the model developed by RaffeSDG. It is worth noting that throughout the study, we maintained a consistent volume of augmented training data by randomly augmenting additional samples.

\begin{figure}[tbp]
\centering
\includegraphics[width=1\linewidth]{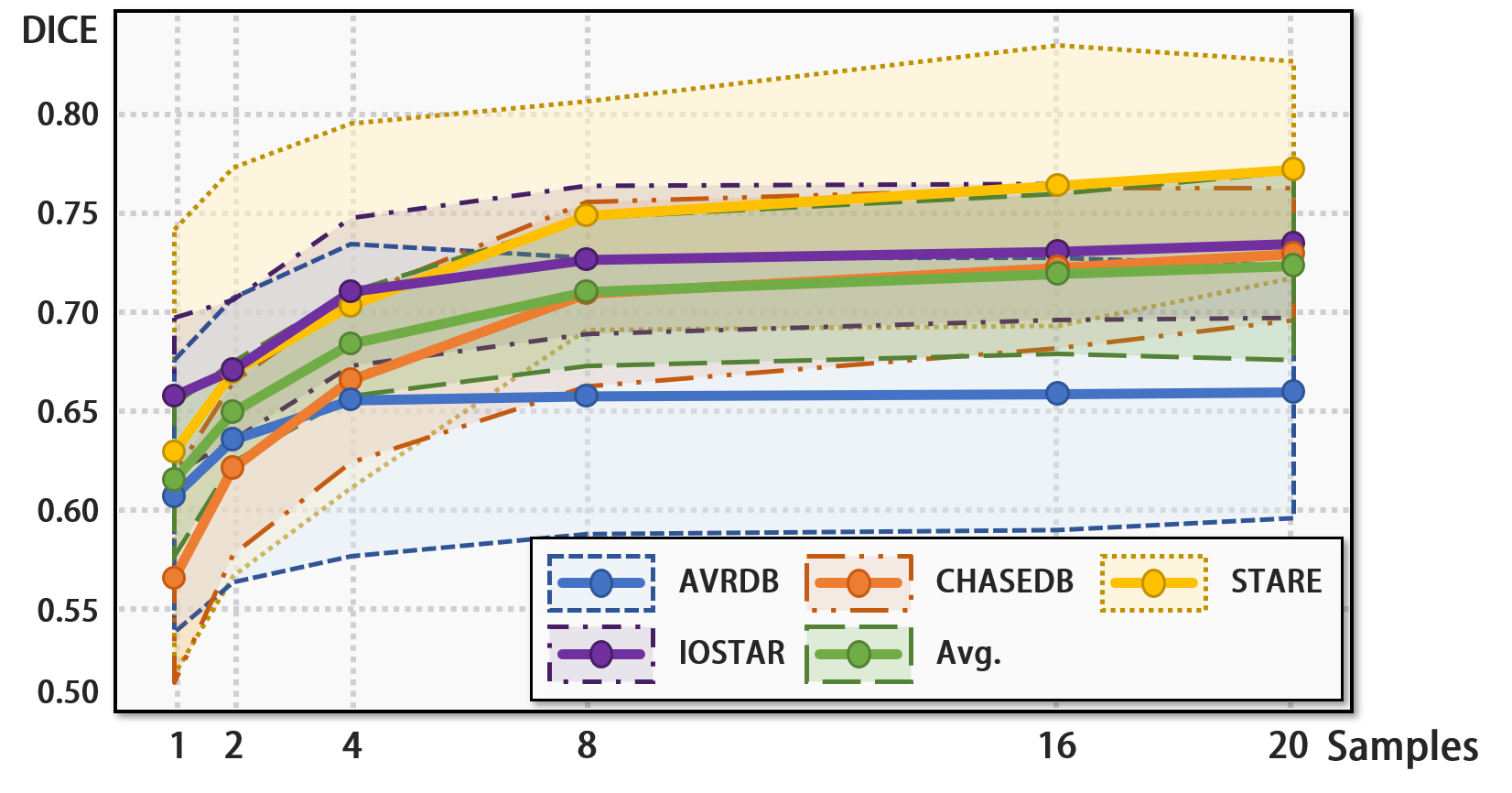}
\caption{
The segmentation performance on out-of-domain data impacted by the data volume from a single-source.
The mean and variance of the predictions on unseen target domains are exhibited.
}
\label{fig:volume}
\end{figure}

The performance achieved by different volumes of raw single-source samples is presented in Fig.~\ref{fig:volume}.
Notable impacts can be observed when the data volume is relatively small (sample size less than 8), as the out-of-domain performance demonstrates significant improvement with increasing sample size. 
However, as the sample size reaches 8, convergence begins to occur, and the rate of performance improvement gradually becomes limited with further increases in sample size.
Furthermore, even with only one raw sample, RaffeSDG allows an average performance of $0.616 \pm 0.038$ in DICE across all target domains, surpassing the performance of several existing SOTA algorithms.

This indicates that RaffeSDG effectively alleviates the burden of extensive data collection, and it consistently delivers strong performance even in few-shot conditions.

\subsubsection{Ablation against segmentor modules}
\begin{figure}[tbp]
\centering
\includegraphics[width= 1\linewidth]{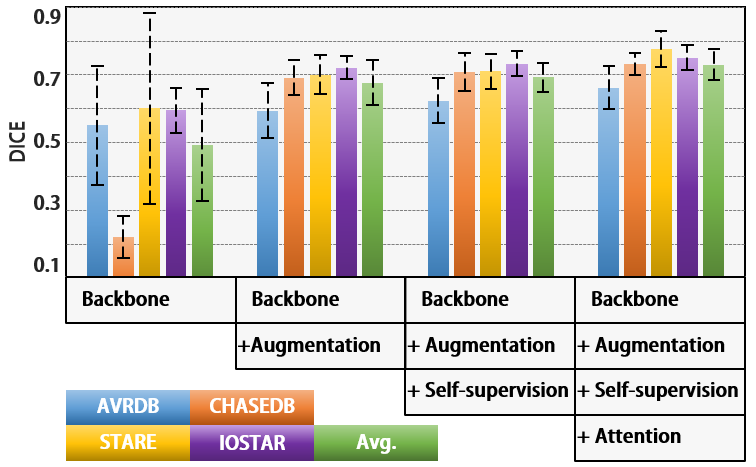}
\caption{
Ablation study of the proposed modules in RaffeSDG.
}
\label{fig:Ablation}
\vskip -10pt
\end{figure}

The effectiveness of the modules in the segmentor of RaffeSDG is verified through ablation studies, as summarized in Figure \ref{fig:Ablation}.
To begin the study, a benchmark is established using only the U-Net backbone to make predictions in unseen target domains from the single-source of DRIVE.
Subsequently, to enhance the diversity of the single-source domain, data augmentation is introduced using the proposed strategy.
Furthermore, the Gaussian-filter-based self-supervision is incorporated to facilitate the learning of domain-invariant representations.
Lastly, attention blocks are implemented to improve the utilization of representations from the pretext task in the segmentation task.

\begin{figure*}[tbp]
\centering
\includegraphics[width=1\linewidth]{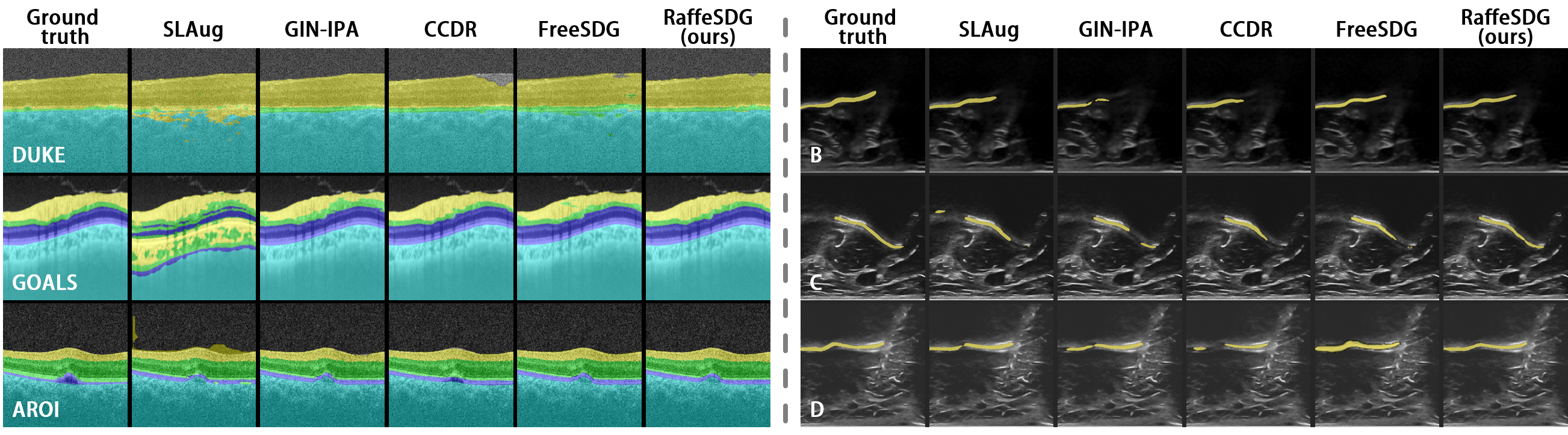}
\caption{
Visualized comparisons with SOTA SDG algorithms for OCT and ultrasound segmentation.  
}
\label{fig:compare2}
\vskip -10pt
\end{figure*}

\begin{table*}[tbp]
\scriptsize
\centering
\caption{Comparisons of OCT and ultrasound segmentation.
The top three results are indicated in \textbf{\color{red}red}, \textbf{\color{magenta}magenta}, and \textbf{\color{blue}blue}.}
\label{tab:comparison_oct} 
\renewcommand{\arraystretch}{1.15}
\begin{tabular}{p{1.9cm} | p{1.2cm}<{\centering} p{1.2cm}<{\centering} p{1.2cm}<{\centering}  p{1.2cm}<{\centering} |p{1.2cm}<{\centering} p{1.2cm}<{\centering} p{1.2cm}<{\centering}  p{1.2cm}<{\centering}}
\hline
\multirow{2}{*}{SDG algorithms} &  \multicolumn{4}{c|}{OCT DICE} &  \multicolumn{4}{c}{Ultrasound DICE}\\
\cline{2-9}
& DUKE & GOALS & AROI & Avg. & B & C & D & Avg.\\ 
\hline
SLAug~\cite{su2022rethinking} & 0.961 & 0.831 & 0.939 & 0.910 & 0.613 & \textbf{\color{magenta}0.639} & \textbf{\color{blue}0.638} & 0.630 \\
GIN-IPA~\cite{ouyang2022causality} & \textbf{\color{red}0.983} & \textbf{\color{red}0.955} & \textbf{\color{blue}0.941} & \textbf{\color{magenta}0.960} & \textbf{\color{blue}0.632} & 0.636 & 0.626 & \textbf{\color{blue}0.631} \\
CCDR~\cite{liang2024single} & \textbf{\color{blue}0.980} & 0.939 & \textbf{\color{magenta}0.951} & \textbf{\color{blue}0.956} & 0.619 & \textbf{\color{blue}0.638} & \textbf{\color{magenta}0.640} & \textbf{\color{magenta}0.632} \\
FreeSDG~\cite{li2023frequency} & 0.978 & \textbf{\color{blue}0.947} & 0.938 & 0.955 & \textbf{\color{red}0.648} & 0.614 & 0.624 & 0.629 \\
RaffeSDG(ours) & \textbf{\color{magenta}0.981} & \textbf{\color{magenta}0.953} & \textbf{\color{red}0.960} & \textbf{\color{red}0.964} & \textbf{\color{magenta}0.644} & \textbf{\color{red}0.663} & \textbf{\color{red}0.658} & \textbf{\color{red}0.655} \\
\hline
\end{tabular}%
\vskip -5pt
\end{table*}

As observed in Fig.~\ref{fig:Ablation}, the introduction of data augmentation significantly enhances out-of-domain segmentation by the backbone module.
Then, the self-supervision based on Gaussian filters imposes additional constraints across augmented samples, enabling the learning of domain-invariant representations within the segmentor.
Afterward, the attention blocks efficiently incorporate the representations learned from the pretext tasks, resulting in a significant enhancement in out-of-domain inference capabilities of the segmentor.

\subsection{Versatility for various tissues and imaging modalities}
Comparisons of out-of-domain inference for OCT and ultrasound image segmentation were also performed to validate the effectiveness of RaffeSDG.
Visualized and quantitative summaries of these comparisons are respectively presented in Fig.~\ref{fig:compare2} and Table~\ref{tab:comparison_oct}.

Considering that CCDR~\cite{liang2024single} and GIN-IPA~\cite{ouyang2022causality} employ specific designs for multi-category labels, it is understandable that they achieve remarkable performance in the retina layer segmentation of OCT, as expressed in Table~\ref{tab:comparison_oct}. 
However, the nature of OCT and ultrasound data poses limitations for RaffeSDG. 
Unlike colorful fundus data obtained through visible light, OCT and ultrasound utilize wave propagation characteristics to capture images of internal structures, resulting in an inherently grayscale appearance. Consequently, OCT and ultrasound do not lend themselves well to the advantages of RaffeSDG, as illustrated in Fig.~\ref{fig:compare2}.
Fortunately, despite these constraints, RaffeSDG still manages to achieve decent results, highlighting its versatility in medical image segmentation.

\section{Discussions}
The scarcity of annotated data is a common challenge in medical scenarios, leading to inevitable domain shifts in clinical settings.
Furthermore, the burdens associated with data collection and computational complexity impede the practicality of many solutions.
Fortunately, SDG focuses on a worst-case scenario of generalization involving out-of-domain inference from a single-source domain, offering an alternative solution to domain shifts.

RaffeSDG presents an efficient SDG paradigm by incorporating variations in frequency space and image blending. 
Through comparisons with common data augmentation techniques, the efficacy of the augmentation strategy devised in RaffeSDG is demonstrated.
Furthermore, an exploration of optimal configurations is conducted to determine the most effective settings.
Subsequently, the superior performance of RaffeSDG is showcased in comparisons with SOTA algorithms, including DA, DG, and SDG methods, in segmentation tasks involving three human tissues captured by four imaging modalities. 
Afterward, through ablation studies, the robustness of RaffeSDG in few-shot scenarios is demonstrated, and the effectiveness of the proposed modules is verified.

Moreover, attributed to the absence of category-wise generalization modules, the multi-category annotations in OCT retina layer segmentation have revealed certain limitations of RaffeSDG. As a result, future revisions will be pursued to incorporate multi-category annotations and address these limitations accordingly.

\section{Conclusions}
Domain shifts detrimentally impact deep learning model performance, and the scarcity of annotated data exacerbates the impact of domain shifts in medical scenarios.
Despite considerable efforts to tackle these challenges, practical implementation in clinical settings is still hindered by limitations in data collection and computational complexity.
To overcome these obstacles, we proposed RaffeSDG, which utilizes frequency variations to impose robust out-of-domain inference based on a single-source domain.
Extensive experiments were executed to interpret the advantages and effectiveness of RaffeSDG. The efficient frequency operations endow RaffeSDG with superiority and efficiency in segmenting out-of-domain data, even when faced with limited training data.

\section*{Acknowledgement}

This work was supported in part by the National Natural Science Foundation of China (62401246), Guangdong Basic and Applied Basic Research Foundation (2514050003407) and Shenzhen Natural Science Fund (JCYJ20240813095112017).

\section*{Declaration of generative AI and AI-assisted technologies in the manuscript preparation process}

During the preparation of this work, we used DeepSeek-R1 in order to optimize textual expression and clarity (including fluency, grammar, word choice consistency, and the readability of the abstract and title). After using this tool/service, we reviewed and edited the content as needed and take full responsibility for the content of the published article. The tool was not used for study design, data collection, data analysis, or the generation of results, figures, or references.


\bibliographystyle{elsarticle-num}
\bibliography{refer}
\end{document}